\documentclass[a4paper,fleqn]{cas-dc}

\usepackage[numbers]{natbib}
\usepackage[ruled,vlined]{algorithm2e}
\usepackage[dvipsnames]{xcolor}

\usepackage{pifont}       
\usepackage{bbding}       
\usepackage{fontawesome}
\definecolor{commentcolor}{RGB}{59,116,116}   
\newcommand{\PyComment}[1]{\ttfamily\textcolor{commentcolor}{\# #1}}  
\newcommand{\PyCode}[1]{\ttfamily\textcolor{black}{#1}} 

\def\tsc#1{\csdef{#1}{\textsc{\lowercase{#1}}\xspace}}
\tsc{WGM}
\tsc{QE}

\begin{document}
\let\WriteBookmarks\relax
\def\floatpagepagefraction{1}
\def\textpagefraction{.001}
\let\printorcid\relax


\shortauthors{}

\title [mode = title]{A Concise but High-performing Network for Image Guided Depth Completion in Autonomous Driving}

%

%

\author[1,2,3]{Moyun Liu}


\ead{lmomoy@hust.edu.cn}
%
%
\affiliation[1]{organization={School of Mechanical Science and Engineering, Huazhong University of Science and Technology},
           city={Wuhan},
           postcode={430074},
           country={China}}
\affiliation[2]{organization={Institute of High Performance Computing (IHPC), Agency for Science, Technology and Research (A*STAR)},
           postcode={138632},
           country={Singapore}}
\affiliation[3]{organization={Centre for Frontier AI Research (CFAR), Agency for Science, Technology and Research (A*STAR)},
           postcode={138632},
           country={Singapore}}

\author[1]{Bing Chen}

\ead{chenbing@hust.edu.cn}
\cormark[1]

\author[1]{Youping Chen}

\ead{ypchen@hust.edu.cn}


\author[1]{Jingming Xie}

\ead{xjmhust@hust.edu.cn}


\author[4]{Lei Yao}

\ead{rayyoh.yao@connect.polyu.hk}

\affiliation[4]{organization={Department of Electrical and Electronic Engineering, The Hong Kong Polytechnic University},
           city={Hong Kong},
           postcode={999077},
           country={China}}


\author[5]{Yang Zhang}

\ead{yzhangcst@hbut.edu.cn}

\affiliation[5]{organization={School of Mechanical Engineering, Hubei University of Technology},
           city={Wuhan},
           postcode={430068},
           country={China}}

\cortext[1]{Corresponding author}

\author[2,3]{Joey Tianyi Zhou}

\ead{zhouty@cfar.a-star.edu.sg}


\begin{abstract}
Depth completion is a crucial task in autonomous driving, aiming to convert a sparse depth map into a dense depth prediction. The sparse depth map serves as a partial reference for the actual depth, and the fusion of RGB images is frequently employed to augment the completion process owing to its inherent richness in semantic information. Image-guided depth completion confronts three principal challenges: 1) the effective fusion of the two modalities; 2) the enhancement of depth information recovery; and 3) the realization of real-time predictive capabilities requisite for practical autonomous driving scenarios. In response to these challenges, we propose a concise but high-performing network, named CHNet, to achieve high-performance depth completion with an elegant and straightforward architecture. Firstly, we use a fast guidance module to fuse the two sensor features, harnessing abundant auxiliary information derived from the color space. Unlike the prevalent complex guidance modules, our approach adopts an intuitive and cost-effective strategy. In addition, we find and analyze the optimization inconsistency problem for observed and unobserved positions. To mitigate this challenge, we introduce a decoupled depth prediction head, tailored to better discern and predict depth values for both valid and invalid positions, incurring minimal additional inference time. Capitalizing on the dual-encoder and single-decoder architecture, the simplicity of CHNet facilitates an optimal balance between accuracy and computational efficiency. In benchmark evaluations on the KITTI depth completion dataset, CHNet demonstrates competitive performance metrics and inference speeds relative to contemporary state-of-the-art methodologies. To assess the generalizability of our approach, we extend our evaluations to the indoor NYUv2 dataset, where CHNet continues to yield impressive outcomes. The code of this work will be available at \url{https://github.com/lmomoy/CHNet}.
\end{abstract}


\maketitle

\section{INTRODUCTION}
\begin{figure}[!t]
  \centering
  \includegraphics[width=0.8\linewidth]{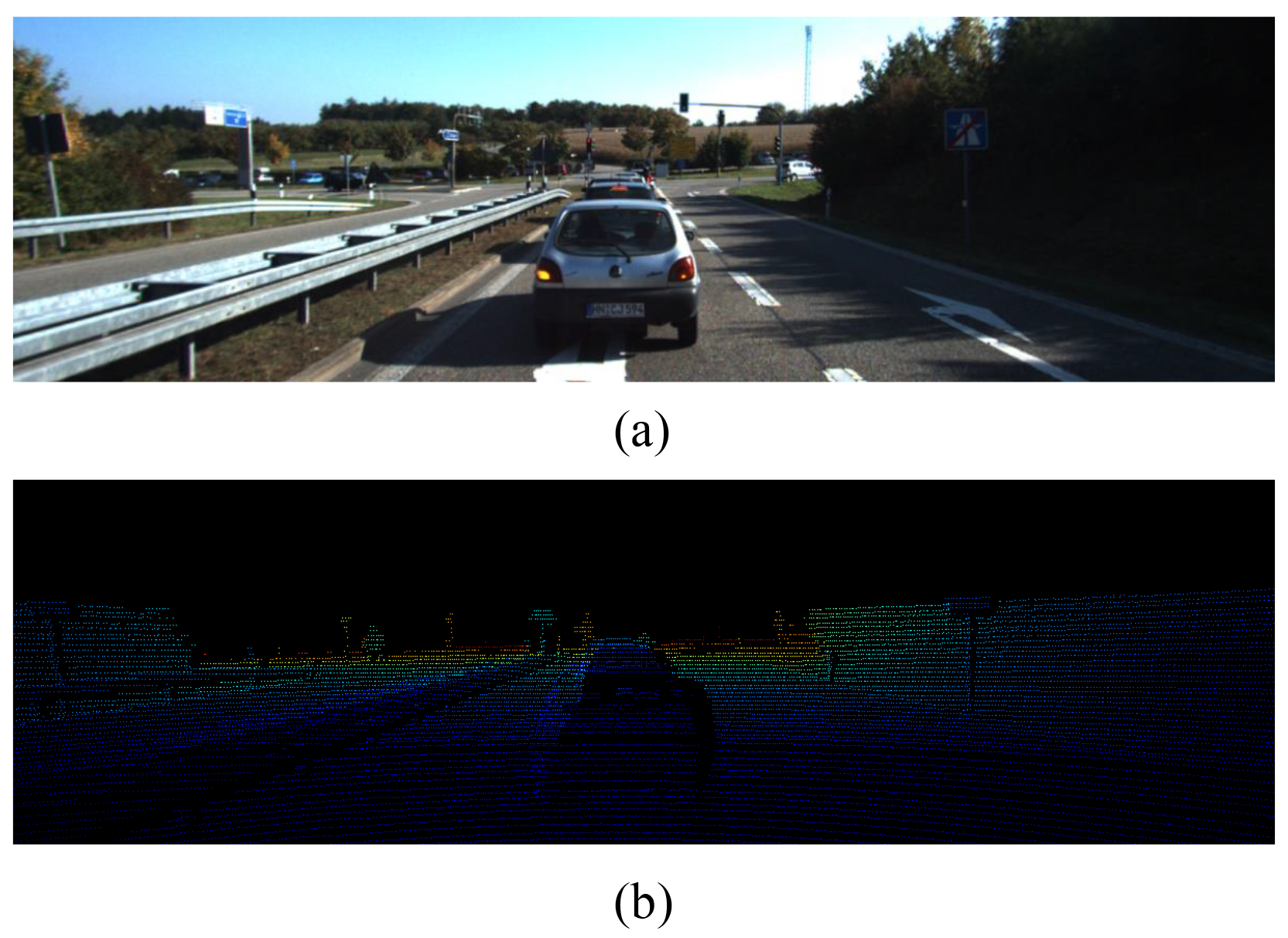}
  \caption{(a) and (b) represent the RGB image and the corresponding sparse depth map, respectively. Compared with the RGB image, the sparse depth map scanned by LiDAR is too sparse to reflect the surroundings.}
  \label{fig:RGB-sparse}
\end{figure}

Perceiving depth is crucial for self-driving cars and mobile robots~\cite{fernandes2021point}, but capturing a dense depth map is challenging due to hardware limitations. LiDAR technology, capable of generating point clouds in three-dimensional space, provides a means to derive depth maps through calibration matrices. Nonetheless, the resultant depth maps often exhibit extreme sparsity, rendering them impractical for direct application, as shown in Fig.~\ref{fig:RGB-sparse}. Relying solely on the sparse depth map itself to fill empty pixel positions is unreliable, as there exist no additional reference cues~\cite{hu2022deep}.

Researchers have acknowledged the synergistic potential between color images from RGB cameras and depth images~\cite{zhou2023bcinet}~\cite{yang2023uplp}~\cite{mosella20212d}, recognizing their capacity to enhance depth completion through their rich semantic content. Consequently, a plethora of methodologies have been proposed to exploit the complementary features of these two modalities in achieving depth completion. However, due to the inherent domain gap between depth and color information, existing approaches that yield superior results often exhibit complexity, sometimes manifesting as multistage processes~\cite{zhao2021adaptive}~\cite{9286883}. To separately extract different modality features, double encoder-decoder structure is used in~\cite{9286883}~\cite{yan2022rignet}. While such architectures tend to enhance domain-specific features, they also incur an increase in parameters, particularly when compared to architectures employing only dual-encoders~\cite{9206740}. Moreover, substantial research effort has been dedicated to the fusion of multiple modalities in the context of depth completion. Drawing inspiration from guidance mechanisms~\cite{6319316}, many contemporary methodologies incorporate effective yet computationally intensive modules to refine depth predictions guided by RGB images. For instance, in~\cite{9286883}, specialized CUDA operators were developed to mitigate the challenge of excessive GPU memory consumption. Additionally, the convolutional spatial propagation network (CSPN)~\cite{cheng2018depth} has emerged as a popular choice for various high-level vision tasks, including depth completion. CSPN acts as a post-processing step, iteratively refining predictions based on local affinities, thereby enhancing prediction accuracy. However, this iterative refinement strategy, while effective, can be inefficient and time-consuming. Furthermore, some approaches integrate 3D geometric cues, such as surface normals or 3D positions~\cite{Qiu2019deeplidar, hu2021penet}, to extract more precise features. While these methods show promise in improving depth estimation accuracy, they often require incorporating prior information or introducing additional learning targets, adding complexity to the model architecture and training process.

\begin{figure}[!t]
  \centering
  \includegraphics[width=1.0\linewidth]{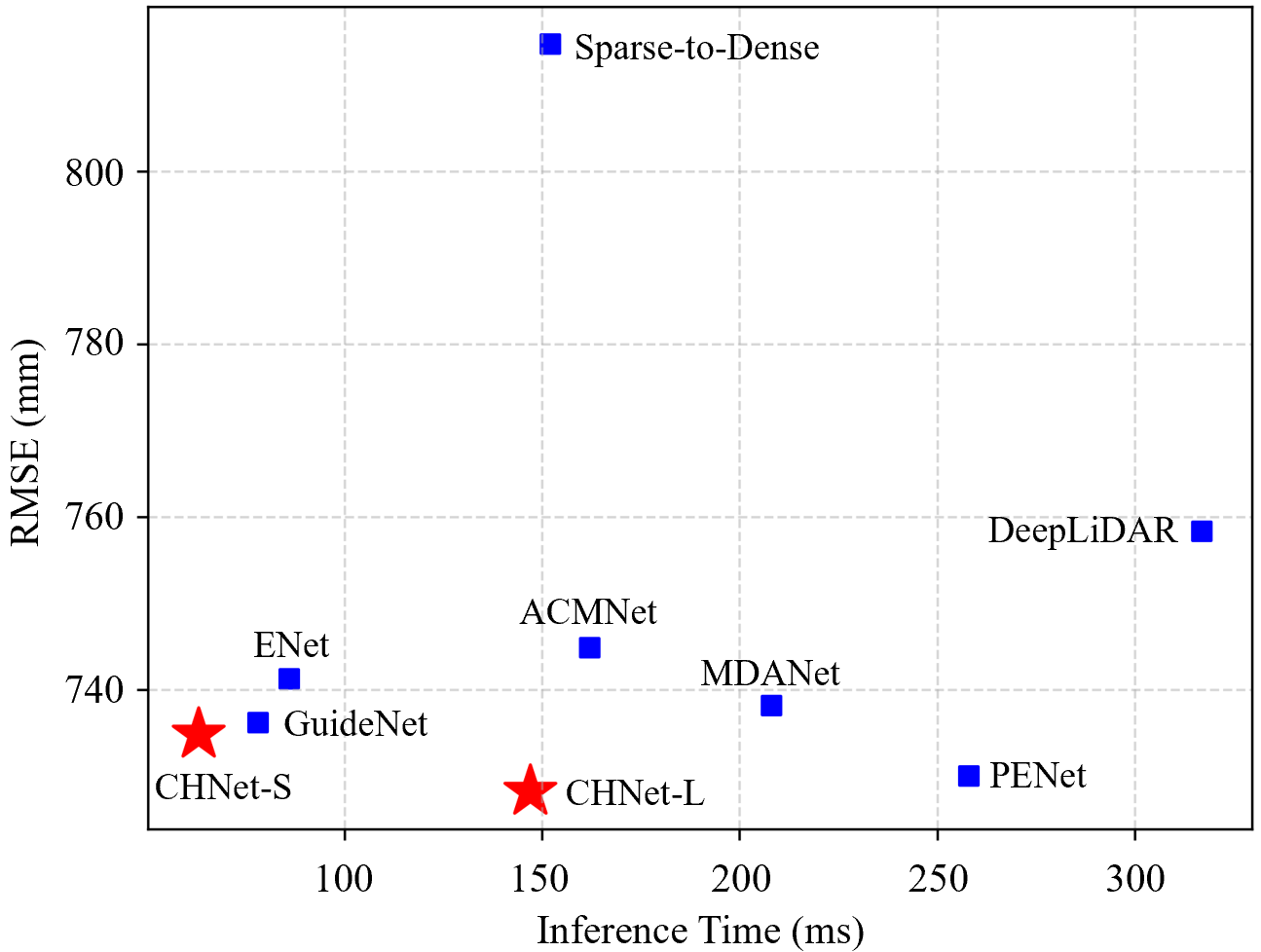}
  \caption{Compared with other methods, which are marked as blue, our CHNet achieves superior performance in terms of accuracy and efficiency on the KITTI depth completion benchmark~\cite{uhrig2017sparsity}. More comparison details can be found in Table~\ref{tab:sota}. Note that the smaller values of root mean squared error (RMSE) and inference time represent better performance.}
  \label{fig:dot}
\end{figure}

In this study, we aim to develop a minimalist network devoid of pre- and post-processing stages while achieving an optimal balance between efficiency and performance. We propose a concise but high-performing network named CHNet, and the comparison of speed and accuracy with state-of-the-art methods is shown in Fig.~\ref{fig:dot}. The overarching structure of CHNet employs a dual-encoder and single-decoder configuration, amalgamating domain-specific feature extraction with architectural simplification. Inspired by prior works\cite{9286883, chen2022guided}, we posit that leveraging semantic features from RGB images can facilitate more accurate recovery of missing pixels within sparse depth maps. For instance, regions exhibiting similar color and texture features are more likely to correspond to the same object, thus suggesting a closer or identical depth value. Rather than employing resource-intensive guidance modules or complex segmentation networks~\cite{el2022guided, nazir2022semattnet}, we propose a fast guidance module. By utilizing channel-wise and integrating cross-channel semantic information, we achieve satisfactory guidance effectiveness. Furthermore, we identify and analyze an optimization inconsistency issue inherent in observed and unobserved positions. To our knowledge, we are the first to scrutinize this aspect of the depth completion task, leading to the development of a decoupled prediction head aimed at mitigating this problem. Through our meticulous network design, CHNet achieves notable performance gains coupled with superior efficiency.

The main contributions of this work can be summarized as:
\begin{itemize}
  \item We introduce a fast guidance module adept at effectively and efficiently fusing sparse depth maps with RGB images. Leveraging semantic information enhances the depth learning process, facilitating more accurate depth completion.
  \item We identify and analyze an optimization inconsistency problem inherent in the depth completion task. Subsequently, we devise a novel decoupled prediction head to address this issue, thereby improving the recovery of depth information.
  \item Our study introduces a concise but high-performing depth completion network based on a simplified architectural structure. This streamlined design facilitates efficient computation while maintaining robust performance.
  \item Through comprehensive evaluations on the KITTI benchmark and NYUv2 dataset, we demonstrate that our CHNet achieves accurate predictions of dense depth maps with notable efficiency compared to state-of-the-art methods.
\end{itemize}

\begin{figure*}[t]
  \centering
  \includegraphics[width=1.0\linewidth]{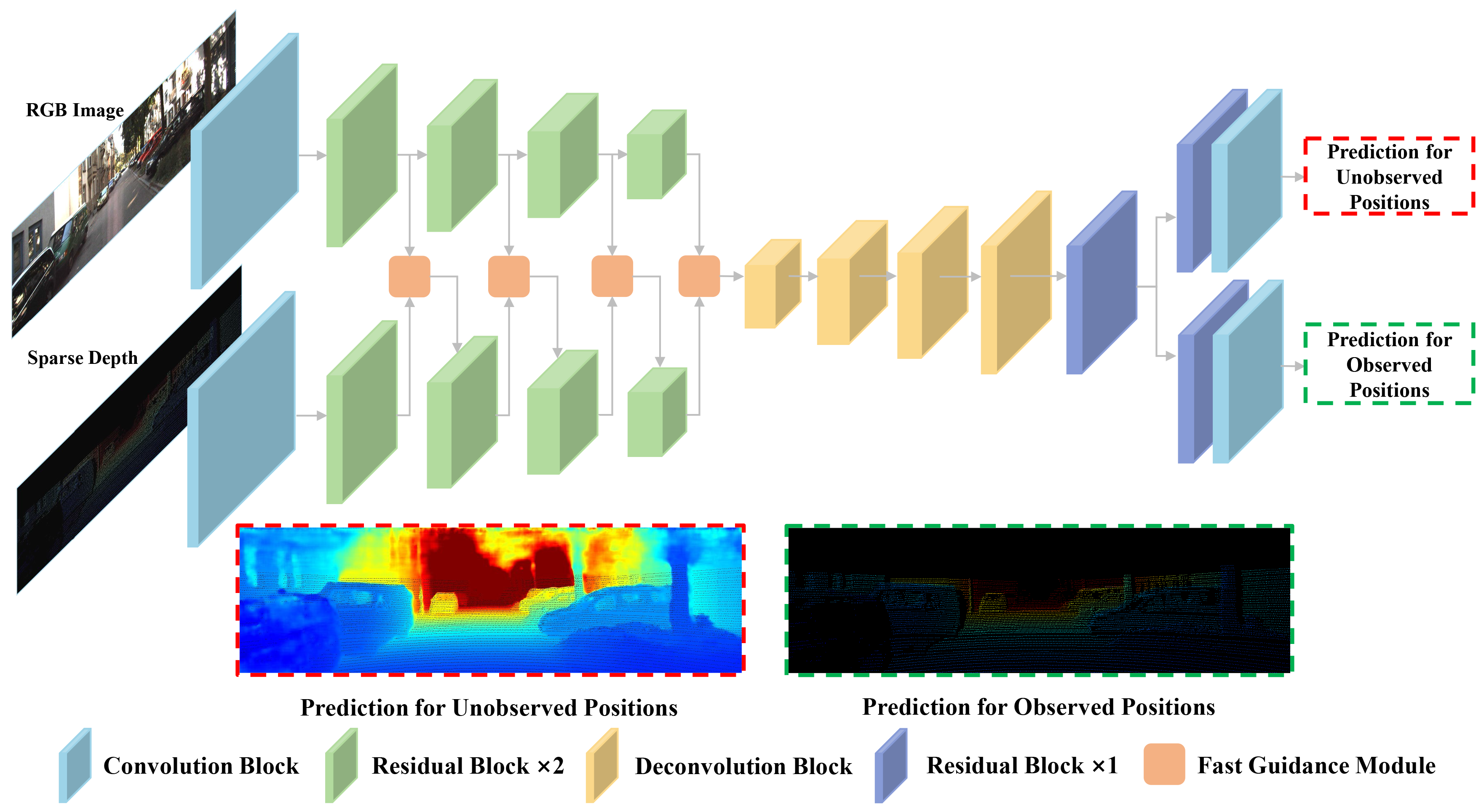}
  \caption{The structure of our CHNet contains two encoders for different modalities and one decoder to output prediction. A decoupled prediction head is connected with the decoder to obtain depth for unobserved and observed positions, respectively. There are skip connections between each output of the depth branch in the encoder and the corresponding features in the decoder, and they are not shown in this figure for simplicity. Different sizes and shapes of blocks correspond to variations in the size and channels of the feature maps they operate on.}
  \label{fig:CHNet}
\end{figure*}

\section{RELATED WORK}
\subsection{Unguided Depth Completion}
Without guidance from RGB or other cues, unguided methods attempt to directly output dense depth map using sparse depth image as input. Uhrig et al. \cite{uhrig2017sparsity} first propose a sparse invariant convolution operation based on a binary validity mask, which is determined by its local neighbors via maxpool, to deal with sparse features. Though it is not readily compatible with classical encoder-decoder networks and validity masks can degrade the model performance \cite{eldesokey2018propagating}\cite{jaritz2018sparse}, many subsequent studies are inspired by it. Motivated by normalized convolution \cite{knutsson1993normalized}, Eldesokey et al. \cite{eldesokey2018propagating} introduce the normalized convolutional neural network which generates a continuous confidence map to tackle this issue. Furthermore, the convolution filters are restricted to be positive due to the SoftPlus function \cite{glorot2011deep} to accelerate convergence. To overcome the absence of semantic cues, some previous work \cite{lu2020depth}\cite{yu2021grayscale} implicitly utilize RGB information by introducing an auxiliary task of depth for reconstruction. Lu et al. \cite{lu2020depth} leverage RGB images as a training objective to encourage the network to learn complementary image features, rather than providing semantic guidance during inference. They simply adopt an sparse depth map as the input and concurrently anticipate a reconstructed image and an dense depth map.

\subsection{RGB Guided Depth Completion}
RGB guided depth completion benefits from the semantic cues from RGB images and achieves more satisfactory prediction result~\cite{hu2022deep}. However, there are more challenges as two modalities are involved. Firstly, the networks usually need to be added specific branch for color image, which increases the complexity of the whole framework. MSG-CHN~\cite{li2020multi} proposes a cascaded hourglass network to merge image features with the corresponding depth features using a specific RGB encoder. Extra encoder and decoder for color image are designed in~\cite{9286883}~\cite{yan2022rignet}, which is for better domain-specific feature extraction. To achieve better coarse-to-fine performance, PENet~\cite{hu2021penet} and FCFRNet~\cite{liu2021fcfr} even utilize several network stages to obtain final depth prediction. Furthermore, the fusion of two modality is also a difficulty due to the domain gap. Simple concatenation~\cite{MaCK19}~\cite{dimitrievski2018learning} is direct but not effective enough. Motivated by~\cite{6319316}, GuideNet~\cite{9286883} introduces the guided convolution network that utilizes RGB image to dynamically generate adaptive spatially-variant kernel weights. In order to refine blurry object boundaries and better recover contextual structures, RigNet~\cite{yan2022rignet} proposes a novel repetitive guidance module built on dynamic convolution~\cite{chen2020dynamic}. CompletionFormer~\cite{zhang2023completionformer} combines convolutional attention and transformer block, achieving superior completion effect with complicated structure.

To obtain more refined results, CSPN~\cite{cheng2018depth} can improve the predicted depth map through adding a propagation network at the tail of model, which can be regarded as a post-processing for further refinement. CSPN++~\cite{cheng2020cspn++} enhances CSPN by improving effectiveness and efficiency by dynamically adjust propagation setup. A dilated and accelerated CSPN++~\cite{hu2021penet} is proposed to enlarge the propagation field and complete propagation in parallel. Despite the high accuracy, such algorithms are not efficient and time-consuming. There are also some works which try to introduce geometric information. DeepLiDAR~\cite{Qiu2019deeplidar} introduces constraints using surface normal, while PENet~\cite{hu2021penet} designs a 3D position map. However, these methods need prior knowledge or extra learning target.

Although the depth completion networks, especially the RGB guided methods have been researched a lot, their designs become more and more complicated. In this paper, we attempt to propose a concise but high-performing network, which can improve the applicability of depth completion.



\section{Method}

To achieve efficient and effective depth completion, we design our network based on dual-encoder and single-decoder structure. Furthermore, we propose a fast guidance module which can produce abundant sub-spaces and global representations. It is used to improve the depth prediction with low-cost operations. In addition, we propose the optimization inconsistency problem for observed and unobserved positions. Accordingly, a solution is proposed and conducted to alleviate it.

\subsection{The Overall Structure of CHNet}
The structure of our CHNet is shown in Fig.~\ref{fig:CHNet}, which is concise and simple. CHNet consists of two encoders for different modalities and one decoder. The raw RGB image and sparse depth map are encoded by separate branches. Firstly, they are processed by a convolution block, which contains a 5$\times$5 convolution, BatchNorm layer, and ReLU layer. Residual blocks~\cite{he2016deep} are further used to extract higher dimension information and downsample feature maps. At each stage, our fast guidance module is inserted to enhance depth feature, and its output will be transferred into the next residual block. Finally, the features are decoded by several deconvolution blocks and recovered to the original resolution. A decoupled prediction head is linked with the decoder, and it predicts depth values for observed and unobserved positions, respectively.

\subsection{Fast Guidance Module}
To introduce our fast guidance module, we firstly review the guidance theory of~\cite{6319316}. Given a input image $I$ and a guidance image $G$, the guidance process can be defined as:

\begin{equation}
E_p=\sum_q W_{p q}(G) I_q,
\end{equation}
\begin{equation}
W_{p q}(G)=\frac{1}{n^2} \sum_{k:(p, q) \in \omega_k}\left(1+\frac{\left(G_p-\mu_k\right)\left(G_q-\mu_k\right)}{\sigma_k^2+\epsilon}\right),
\end{equation}
while $E$ is the enhanced output image, $W_{p q}$ is a filter kernel, $p$ and $q$ represent the different indexes of pixels. In~\cite{6319316}, it explicitly define the filter kernel using the statistic within a window $\omega_k$ whose size is $n \times n$, while $\mu_k$ and $\sigma_k^2$ are the mean and variance in $\omega_k$.

\begin{figure}[t]
  \centering
  \includegraphics[width=1.0\linewidth]{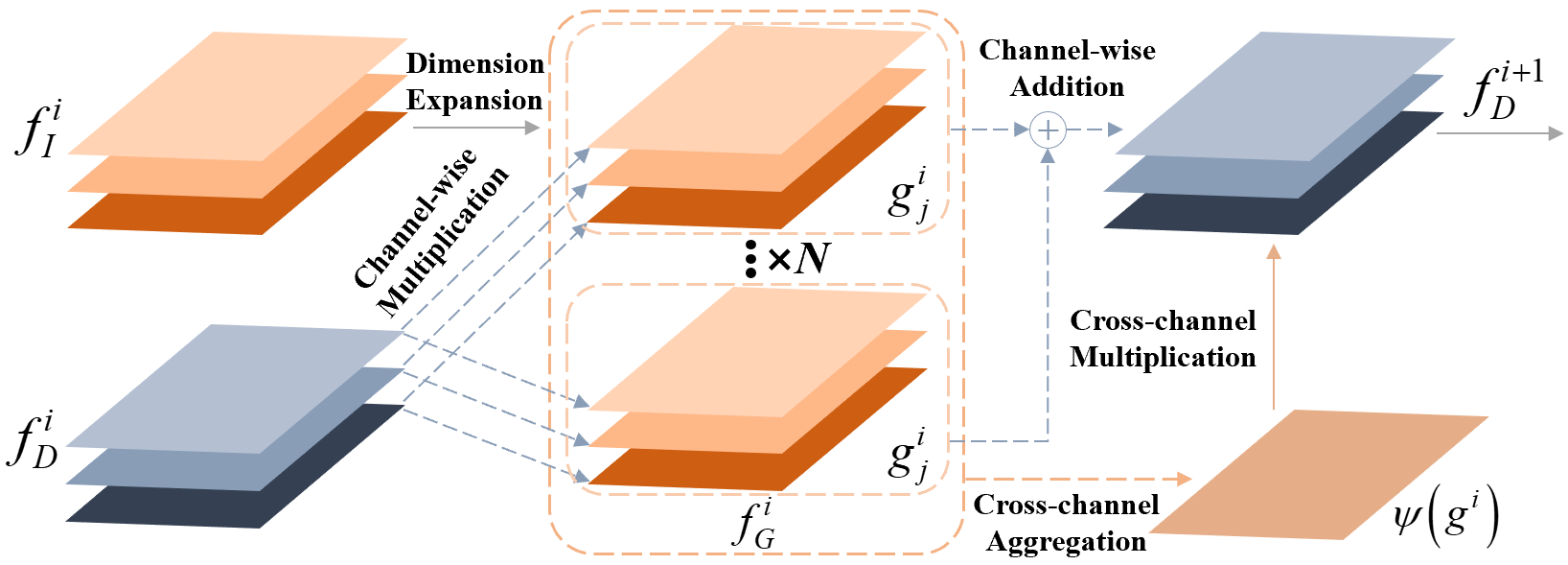}
  \caption{The illustration of our fast guidance module.}
  \label{fig:FG}
\end{figure}

The segmentation boundary within local windows is used to help depth estimation~\cite{jung2021fine}. Inspired by it~\cite{jung2021fine}, we believe that the semantic information from color image can contribute to better depth learning within local region. Instead of introducing extra segmentation network~\cite{el2022guided}~\cite{nazir2022semattnet} or designing time-consuming module~\cite{9286883}, a fast guidance module is proposed in this paper, and we use a $3 \times 3$ convolution to explore the guidance information within local window. As shown in Fig.~\ref{fig:FG}, we denote the $f_I^i \in \mathbb{R}^{H \times W \times C}$ and $f_D^i \in \mathbb{R}^{H \times W \times C}$ as the feature maps from image and sparse depth streams at stage $i$, respectively. Firstly, we apply a $3 \times 3$ convolution on the $f_I^i$, which is used to integrate the local information of image feature as the guidance feature, then the channel dimension is expanded through a $1 \times 1$ convolution to obtain $f_G^i \in \mathbb{R}^{H \times W \times NC}$. Because different channels reflect various semantic information for the scene, we utilize dimension expansion to generate more feature sub-spaces. Then, we split $f_G^i$ into $N$ parts along channel dimension and obtain sub-space $g_j^i (j = 1, \ldots, N)$. $f_D^i$ is used to apply channel-wise multiplication with $g_j^i$, and we integrate each sub-space by simple addition operation at last. In addition, we aggregate cross-channel features and compress the data, and its goal is to explore global representation of the whole feature space. We also leverage the global representation to further guide and refine our depth generation. Our fast guidance module can be formulated as:

\begin{equation}
f_D^{i+1}= \psi(g^i) \odot \sum_{j=1}^N\left(f_D^i \odot g_j^i\right),
\end{equation}
where $\psi$ is cross-channel aggregation operation, and we use average calculation in this paper. The $\odot$ operator represents channel-wise multiplication. Because there are no attention mechanisms or local operations, our guidance module can be very efficient, and its runtime will be discussed in Section IV-D.

Different from~\cite{6319316}, we obtain the guidance map using implicit feature extraction, and the expansion and aggregation operations fully exploit potential of guidance weights. The most similar method to our fast guidance module may be~\cite{9286883}, but there are many differences between them. Firstly, ~\cite{9286883} uses both of image and LiDAR features to generate guidance weights, while ours only uses image feature. Then, ~\cite{9286883} has two weights obtained by different layers and used for channel-wise and cross-channel guidance. Our module also has two guidance stages, but the second guidance weight is aggregated from the first stage. The biggest difference between them is the guidance strategy,~\cite{9286883} achieves guidance mainly through generating different guided kernels for each pixel. Hence, there are lots of local operations, so that CUDA operators have to be implemented for acceleration. The guidance elements of our method are the whole feature maps, and only simple channel-wise multiplication is needed to obtain results. We compare the efficiency and effectiveness of our fast guidance module and~\cite{9286883} in Section 4.5, which proves the superiority of our method. We provide simple pseudocode of our fast guidance module with PyTorch style, as shown in Algorithm~\ref{algo:code}.

\begin{algorithm}[t]
\normalsize
\SetAlgoLined
    \PyComment{$w$: convolutional layer with same input and output dimension} \\
    \PyComment{$w_{r}$: convolutional layer to increase the channel dimension} \\
    \PyComment{r: expansion ratio} \\
    \PyComment{$f_{i}$: image feature} \\
    \PyComment{$f_{d}$: LiDAR feature} \\
    \PyCode{} \\
    \PyComment{Obtain guidance weight} \\
    \PyCode{weight = $w(f_{i})$} \\
    \PyComment{Expand weight} \\
    \PyCode{weight = $w_{r}$(weight)} \\
    \PyComment{Channel-wise guidance} \\
    \PyCode{kernels = torch.chunk(weight, r, 1)} \\
    \PyCode{splits = []} \\
    \PyCode{for i in range(r):} \\
    \Indp
    \PyCode{splits.append($f_{d}$ * kernels[i])} \\
    \Indm
    \PyCode{out = sum(splits)} \\
    \PyComment{Cross-channel guidance} \\
    \PyCode{avg\_weight = torch.mean(weight, dim=1)} \\
    \PyCode{out = $w$(out * avg\_weight)}  \\
\caption{An example of PyTorch-style pseudocode for fast guidance module.}
\label{algo:code}
\end{algorithm}

\subsection{Decoupled Prediction Head}

\begin{figure}[t]
  \centering
  \includegraphics[width=1.0\linewidth]{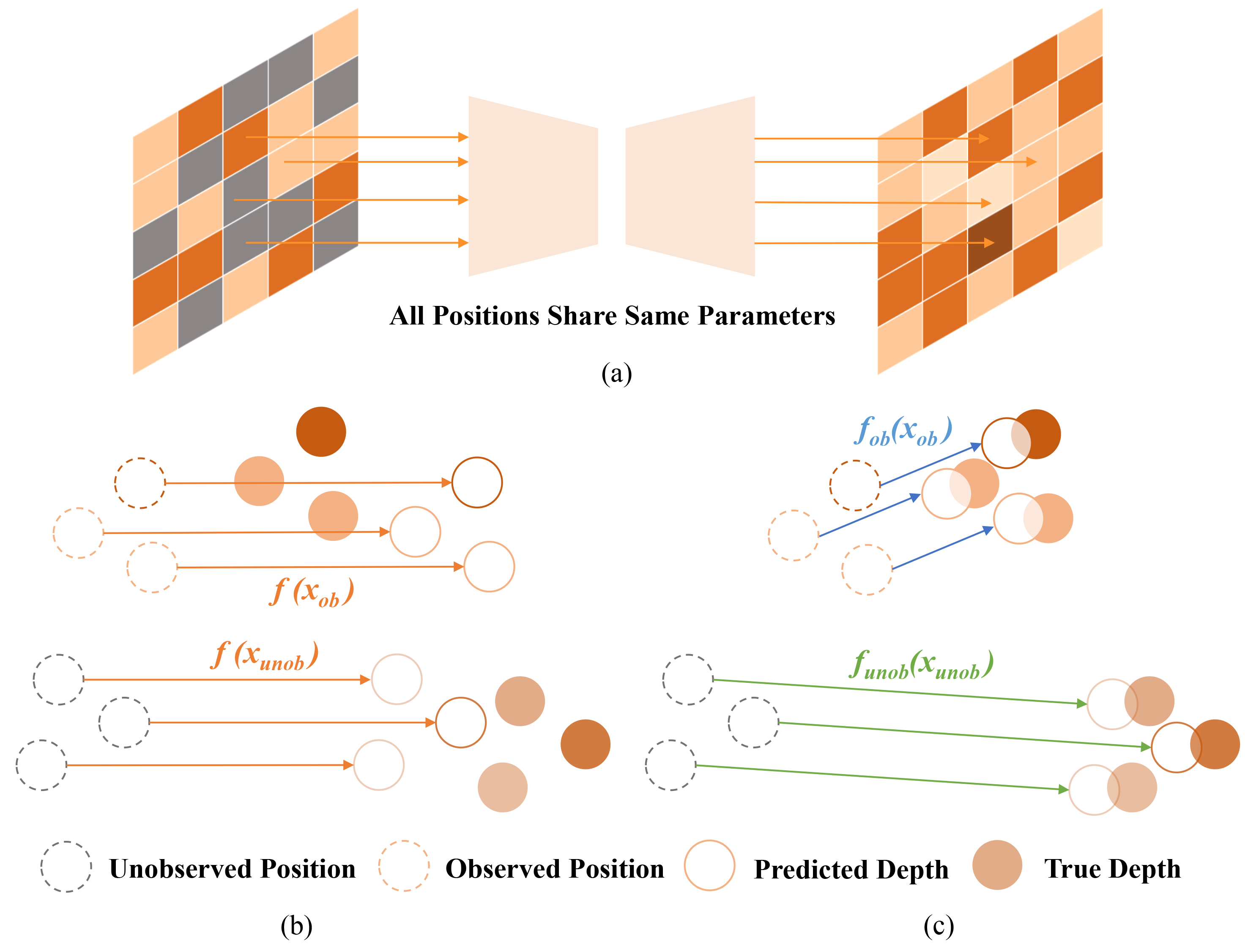}
  \caption{The illustration of optimization inconsistency problem for unobserved and observed positions. We use gray color denotes the unobserved positions, and the degree of orange color represents the depth value. (a) shows the traditional networks assign same parameters to all positions, whether they are observed or not. (b) reveals the optimization inconsistency will incur suboptimal prediction result. (c) is used to prove the effectiveness of decoupled prediction head.}
  \label{fig:OI}
\end{figure}

For the input images of depth completion task, we can classify the pixel positions into observed and unobserved position according to the mask of sparse depth map. We denote the all positions as $X = \{x^z\}_{z=1, \ldots, k}$, the observed positions as $X_{ob} = \{x_{ob}^z\}_{z=1, \ldots, m}$, and the unobserved positions as $X_{unob} = \{x_{unob}^z\}_{z=1, \ldots, n}$, while $k$, $n$, and $m$ are the corresponding numbers, respectively. Hence, $X = X_{ob} \bigcup X_{unob}$, and $k = m + n$. For the general depth completion network, $x_{ob}^z$ and $x_{unob}^z$ are shared the same parameters, as shown in Fig.~\ref{fig:OI} (a). In the training stage, they are also optimized with equal treatment. However, we find that there is an optimization inconsistency problem under such operation, as shown in Fig.~\ref{fig:OI} (b). We denote the whole model as a function $f(\cdot)$, and the mapping of the observed and unobserved position are $f_{ob}(\cdot)$ and $f_{unob}(\cdot)$, respectively. For the common network, $f_{ob}(\cdot) = f_{unob}(\cdot) = f(\cdot)$, due to the parameter sharing~\cite{dai2020parameters}. Whereas, the distance between $x_{ob}$ and $gt_{ob}$ is far closer than $x_{unob}$ and $gt_{unob}$ in feature space, while $gt_{ob}$ and $gt_{unob}$ are the corresponding true depth of the observed and unobserved position, respectively. In general, $x_{unob} = 0$ and $gt_{unob} > 0$, while $x_{ob} \approx gt_{ob}$. (Note that the depth value of the observed position in the sparse depth map is not exactly equal to the value of the corresponding position in ground truth. Because the ground truth is usually obtained by accumulating several frames~\cite{uhrig2017sparsity}, which is more precise.) Hence, the training optimization for the two positions will influence each other, resulting in sub-optimal prediction performance.

To address this issue, we propose the assignment of distinct parameters for $x_{ob}$ and $x_{unob}$, thereby facilitating the establishment of two distinct mappings, denoted as $f_{ob}(\cdot)$ and $f_{unob}(\cdot)$, as depicted in Fig.~\ref{fig:OI} (c). Notably, $f_{ob}(\cdot)$ and $f_{unob}(\cdot)$ operate on $x_{ob}$ and $x_{unob}$, respectively. By virtue of being endowed with disparate learnable parameters, $f_{ob}(\cdot)$ and $f_{unob}(\cdot)$ are empowered to acquire distinct mappings characterized by varying directions and distances. This attribute enables a more precise alignment with the requirements of corresponding positions. Through the employment of diverse optimization strategies, the iterative update process enables the two positions to converge more effectively towards their true values. To be specific, we design a decoupled prediction head, and predict two depth maps $P_{1}$ and $P_{2}$. The two depth maps are full resolution, but we set a mask to prevent some of their pixels from participating in back propagation, and the mask $M$ can be formulated as:

\begin{equation}
S = \{(u,v) \mid D_{u,v}^{inp} \textgreater 0\},
\end{equation}

\begin{equation}
M_{u,v} = \begin{cases} 1, & (u,v) \in S \\ 0, & (u,v) \notin S \end{cases},
\end{equation}
where $D^{inp}$ is the input of sparse depth map, $u$ and $v$ refer to the indices of a specific pixel position, and S is a set containing all pixels whose values are larger than 0.

Finally, the decoupled prediction outputs can be represented as follows:
\begin{equation}
P_{ob} = P_{1} \odot M,
\end{equation}
\begin{equation}
P_{unob} = P_{2} \odot (1 - M),
\end{equation}
where $P_{unob}$ and $P_{ob}$ are used to optimize the network based on the unobserved and observed positions, respectively. Under the setting of decoupled prediction head, we give different transformation space for the observed and unobserved positions. They can get rid of the balance optimization and pursue independent optimization.
\subsection{Loss Function}
To calculate the difference between the prediction and ground truth $D^{gt}$, we utilize the $\mathcal{L}_2$ loss function. Before the loss computation, we integrate the decoupled output to obtain the completed prediction $D^P$ as follows:

\begin{equation}
D^P = P_{unob} \bigcup P_{ob}.
\end{equation}

Due to the ground truth is semi-dense, we only calculate the pixels difference between
$D^P$ and $D^{gt}$ in valid positions, and the loss function of our network can be written as:

\begin{equation}
\mathcal{L} = \frac{1}{\left\| D^{gt} > 0 \right\|_0} \left\| \left(D^P - D^{gt}\right) \odot \left[D^{gt} > 0\right] \right\|_F^2.
\end{equation}

\section{Experiments}
In this section, we first introduce the dataset, evaluation metrics, and implementation details. Then, we present the quantitative and qualitative comparison with other state-of-the-art methods. In addition, ablation studies are conducted to verify the design of the proposed network.
\subsection{Datasets}
The KITTI depth completion dataset~\cite{uhrig2017sparsity} consists of the aligned sparse depth maps and high-resolution color images. It contains 86,898 frames for training, 7,000 frames for validation, and another 1,000 frames for testing. We use the officially selected 1,000 frames for validation. The dataset provides a public leaderboard for evaluation, and the ground truth depth maps are obtained by aggregating LiDAR scans from 11 consecutive frames.

To validate the generalization of our CHNet, we also test our model on indoor NYUv2 depth dataset~\cite{silberman2012indoor}. NYUv2 dataset is built from 464 indoor scenes and owns RGB and depth images through Microsoft Kinect. We follow the setting in~\cite{zhang2023completionformer}, and the input images are resized from 640$\times$480 to 320$\times$240, and then further center-cropped to 304$\times$228. We train our model on 50K images and test on 654 images from the official labeled test set. Because the depth map in NYUv2 is dense, we randomly sample 500 pixels to generate sparse depth map.

\subsection{Evaluation Metrics}
We follow the standard evaluation metrics for the KITTI depth completion dataset, which includes root mean squared error (RMSE), mean absolute error (MAE), root mean squared error of the inverse depth (iRMSE), and mean absolute error of the inverse depth (iMAE). RMSE and MAE directly measure depth accuracy, while iRMSE and iMAE compute the mean error of inverse depth. The dominant metric for ranking submissions on the KITTI leaderboard is RMSE. Note that the smaller the value of these four evaluation metrics, the better the performance of the model. Additionally, the runtime of inference is also reported for performance comparison to show the efficiency of our method. For NYUv2 dataset, we also use RMSE as the metric. In addition, we introduce mean absolute relative error (REL), and $\delta_i(i=1.25,1.25^2, 1.25^3)$.

\subsection{Implementation Details}
In our implementation, we use the Adam optimizer~\cite{kingma2014adam} with an initial learning rate of 0.001, weight decay of $1\times10^{-6}$ and the momentum of $\mathrm{\beta}_1$ = 0.9, $\mathrm{\beta}_2$ = 0.99. We train our model for 25 epochs with batch size of 8 and use a stepwise learning rate schedule, where the learning rate is reduced by a factor of 0.5, 0.1, and 0.01 at the 10th, 15th, and 20th epoch, respectively. In terms of the inference speed, we use CUDA's asynchronous execution on CPU and GPU for precise measurement, and the test platform is a single 3060 GPU.

In this paper, we propose two models with different parameters, e.g., CHNet-S and CHNet-L. The former is a more lightweight version, and the CHNet-L pursues higher performance. The convolution block and the first three residual block stages in the encoder increase the input dimension twice and reduce the resolution twice as well. The last residual module maintains the feature map dimension but still downsamples. For CHNet-S, the first convolution block increases the dimensions of both inputs to 32, while 64 for CHNet-L.

\subsection{Comparisons with the State-of-the-art Models}
\begin{table*}[!t]
	\renewcommand{\arraystretch}{1.4}
	\renewcommand\tabcolsep{2pt}
	\caption{Comparisons to state-of-the-art methods on the KITTI test set, ranked by RMSE. (*) denotes speeds sourced from the KITTI online leaderboard, with these methods not publicly releasing their code.}
    \centering
    \begin{tabular}{l|c c c c|c c c}
        \toprule
        Method  & \textbf{RMSE}~$\downarrow$ & MAE~$\downarrow$ & iRMSE~$\downarrow$ & iMAE~$\downarrow$ & Params(M)~$\downarrow$ & FLOPs(G)~$\downarrow$ & Speed~$\downarrow$ \\
        \midrule
        CSPN~\cite{cheng2018depth} & 1019.64 & 279.46 & 2.93 & 1.15 & - & - & 1.00s*\\
        Sparse-to-Dense~\cite{MaCK19} & 814.73 & 249.95 & 2.80 & 1.21 & - & - & 0.08s*\\
        PwP~\cite{xu2019depth} & 777.05 & 235.17 & 2.42 & 1.13 & - & - &0.10s* \\
        GAENet~\cite{chen2022depth} & 773.90 & 231.29 & 2.29 & 1.08 & - & - &0.05s* \\
        DSPN~\cite{xu2020deformable} & 766.74 & 220.36 & 2.47 & 1.03 & - & - & 0.34s*\\
        ABCD~\cite{jeon2021abcd} & 764.61 & 220.86 & 2.29 & 0.97 & - & - & 0.02s*\\
        MSG-CHN~\cite{li2020multi} & 762.19 & 220.41 & 2.30 & 0.98 & - & - & 0.01s* \\
        DeepLiDAR~\cite{Qiu2019deeplidar} & 758.38 & 226.50 & 2.56 & 1.15 & 143.98 & 1537.91 & 0.32s \\
        UberATG~\cite{chen2019learning} & 752.88 & 221.19 & 2.34 & 1.14 & - & - & 0.09s* \\
        ACMNet~\cite{zhao2021adaptive} & 744.91 & 206.09 & 2.08 & 0.90 & 4.94 & 272.57 & 0.16s \\
        CSPN++~\cite{cheng2020cspn++} & 743.69 & 209.28 & 2.07 & 0.90 & - & - & 0.20s* \\
        ENet~\cite{hu2021penet} & 741.30 & 216.26 & 2.14 & 0.95 & 131.76 & 374.53 & 0.09s \\
        CDCNet~\cite{fan2022cascade} & 738.26 & 216.05 & 2.18 & 0.99 & - & - & 0.06s* \\
        MDANet~\cite{ke2021mdanet} & 738.23 & 214.99 & 2.12 & 0.99 & 3.05 & 225.39 & 0.21s\\
        GuideNet~\cite{9286883} & 736.24 & 218.83 & 2.25 & 0.99 & 62.60 & 229.95 & 0.08s \\
        FCFRNet~\cite{liu2021fcfr} & 735.81 & 217.15 & 2.20 & 0.98 & - & - &-\\
        PENet~\cite{hu2021penet} & 730.08 & 210.55 & 2.17 & 0.94 & 131.91 & 405.82 & 0.26s \\
        ReDC~\cite{sun2023revisiting} & 728.31 & 204.60 & 2.05 & 0.89 & 131.78 & 381.23 & 0.19s\\
        MMF-Net~\cite{liu2023mff} & 719.85 & 208.11 & 2.21 & 0.94 & - & - &-\\
        \midrule
        CHNet-S (Ours)& 734.83 & 213.48 & 2.20 & 0.95 & 14.75 & 177.68 & 0.06s\\
        CHNet-L (Ours)& 728.23 & 209.89 & 2.16 & 0.93 & 58.65 & 607.53 & 0.15s\\
        \bottomrule
    \end{tabular}
	\label{tab:sota}
\end{table*}

\begin{table}[!t]
	\renewcommand{\arraystretch}{1.4}
	\renewcommand\tabcolsep{1pt}
	\caption{Comparisons to state-of-the-art methods on the NYUv2 dataset, ranked by RMSE.}
    \centering
    \begin{tabular}{l|c c c c c}
        \toprule
        Method  & \textbf{RMSE}~$\downarrow$ & REL~$\downarrow$ & $\delta_{1.25}$~$\uparrow$ & $\delta_{1.25^2}$~$\uparrow$ & $\delta_{1.25^3}$~$\uparrow$ \\
        \midrule
        Bilateral \cite{silberman2012indoor}     & 0.479 & 0.084 & 92.4 & 97.6 & 98.9\\
        Zhang \cite{zhang2018deep} & 0.228 & 0.042 & 97.1 & 99.3 & 99.7 \\
        Sparse-to-Dense~\cite{MaCK19}      & 0.230    & 0.044   & 97.1  & 99.4  & 99.8 \\
        DCoeff \cite{imran2019depth} & 0.118    & 0.013   & 99.4  & 99.9  & -   \\
        CSPN \cite{cheng2018depth}     & 0.117    & 0.016   & 99.2  & 99.9  & 100.0 \\
        CSPN++ \cite{cheng2020cspn++}      & 0.116    & -       & -     & -     & -    \\
        DeepLiDAR~\cite{Qiu2019deeplidar}   & 0.115    & 0.022   & 99.3  & 99.9  & 100.0\\
        PwP~\cite{xu2019depth}   & 0.112    & 0.018   & 99.5  & 99.9  & 100.0  \\
        FCFRNet~\cite{liu2021fcfr}       & 0.106    & 0.015   & 99.5  & 99.9  & 100.0 \\
        ACMNet~\cite{zhao2021adaptive}   & 0.105    & 0.015   & 99.4  & 99.9  & 100.0\\
        PRNet \cite{lee2021depth}        & 0.104    & 0.014   & 99.4  & 99.9  & 100.0 \\
        GuideNet \cite{9286883} & 0.101    & 0.015   & 99.5  & 99.9  & 100.0 \\
        \midrule
        CHNet-S (Ours)& 0.099 & 0.016 & 99.5 & 99.9 & 100.0\\
        \bottomrule
    \end{tabular}
	\label{tab:sota-NYU}
\end{table}

\begin{table}[!t]
	\renewcommand{\arraystretch}{1.4}
	\renewcommand\tabcolsep{4pt}
	\caption{The complexity comparison between our fast guidance module and guide module in~\cite{9286883}}
	\centering
	\label{tab:guidance_comparison}
	\begin{tabular}{c|cc}
		\toprule
		                    & Fast Guidance  Module & Guide Module in~\cite{9286883} \\
		\midrule
         Speed              & 0.011s           & 0.026s   \\
         Parameters         & 0.36M            & 4.18M   \\
         FLOPs              & 17.63G           & 100.50G   \\
        \bottomrule
	\end{tabular}
\end{table}

\begin{figure}[!t]
  \centering
  \includegraphics[width=1.0\linewidth]{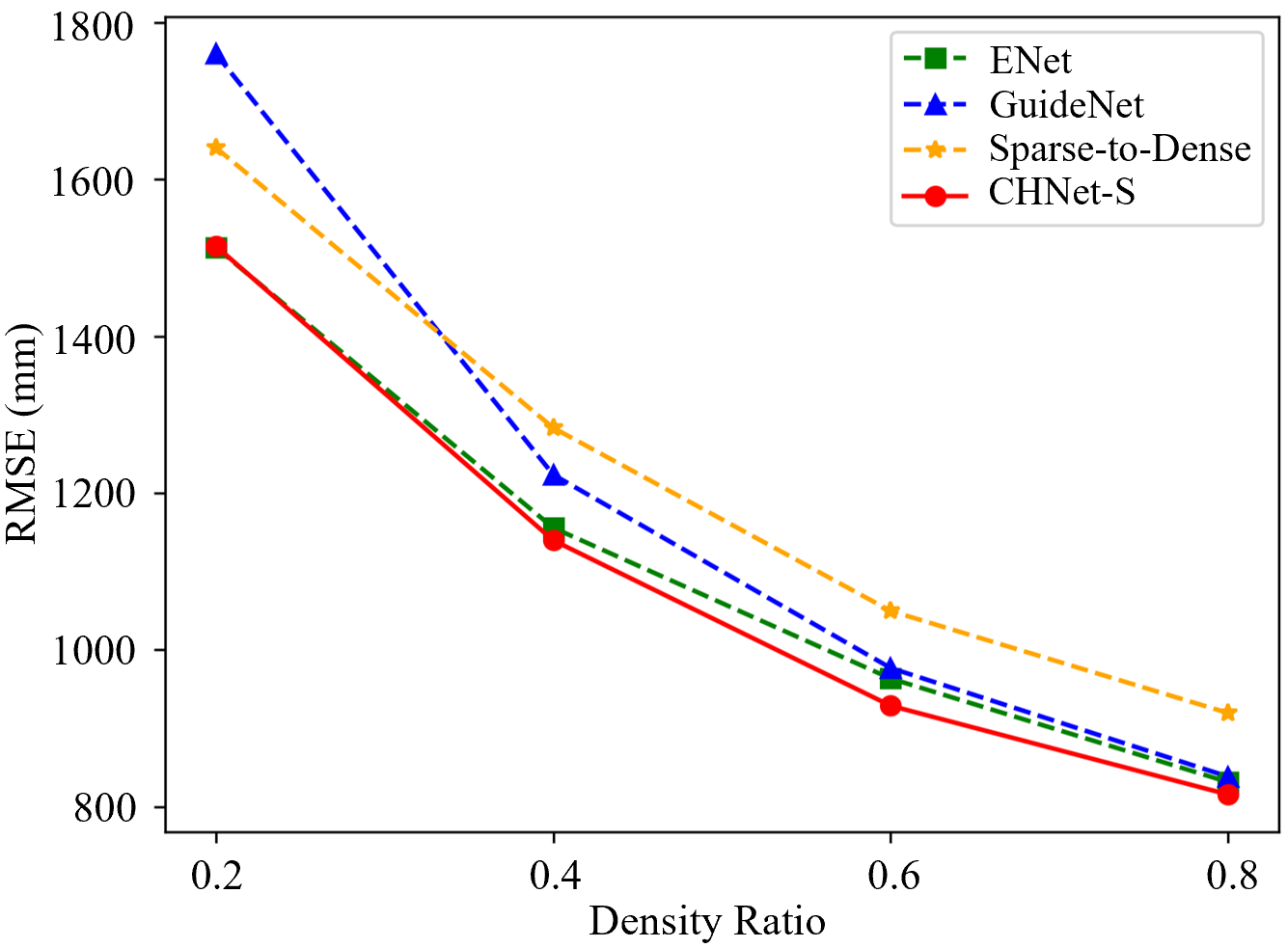}
  \caption{The performance comparison of our CHNet-S and other state-of-the-art methods under different density ratios of input LiDAR point cloud.}
  \label{fig:DR}
\end{figure}

We evaluate our proposed methods on the KITTI depth completion benchmark and compare their performance with state-of-the-art methods, as shown in Table~\ref{tab:sota}. In terms of RMSE, the proposed CHNet-S and CHNet-L methods achieve 734.83 mm and 728.23 mm, respectively. GuideNet~\cite{9286883} and ENet~\cite{hu2021penet} are efficient with relatively fast prediction speeds, but they are still slower than our CHNet-S, while CHNet-S has higher accuracy compared with them. As a bigger version, our CHNet-L is still faster than most of the top-ranking methods and achieves more superior prediction accuracy. The results on the KITTI benchmark prove the effectiveness and usability of our network. The MMF-Net~\cite{liu2023mff} outperforms our approach in RMSE, leveraging extensive attention operations but potentially sacrificing efficiency. While our CHNet-L surpasses ReDC in RMSE but lags in MAE due to its error-reducing post-processing. The improvements in iRMSE and iMAE metrics are modest compared to RMSE and MAE, possibly due to the denser point clouds in nearer regions with more original depth values from LiDAR data.

\begin{figure*}[!t]
  \centering
  \includegraphics[width=0.95\linewidth]{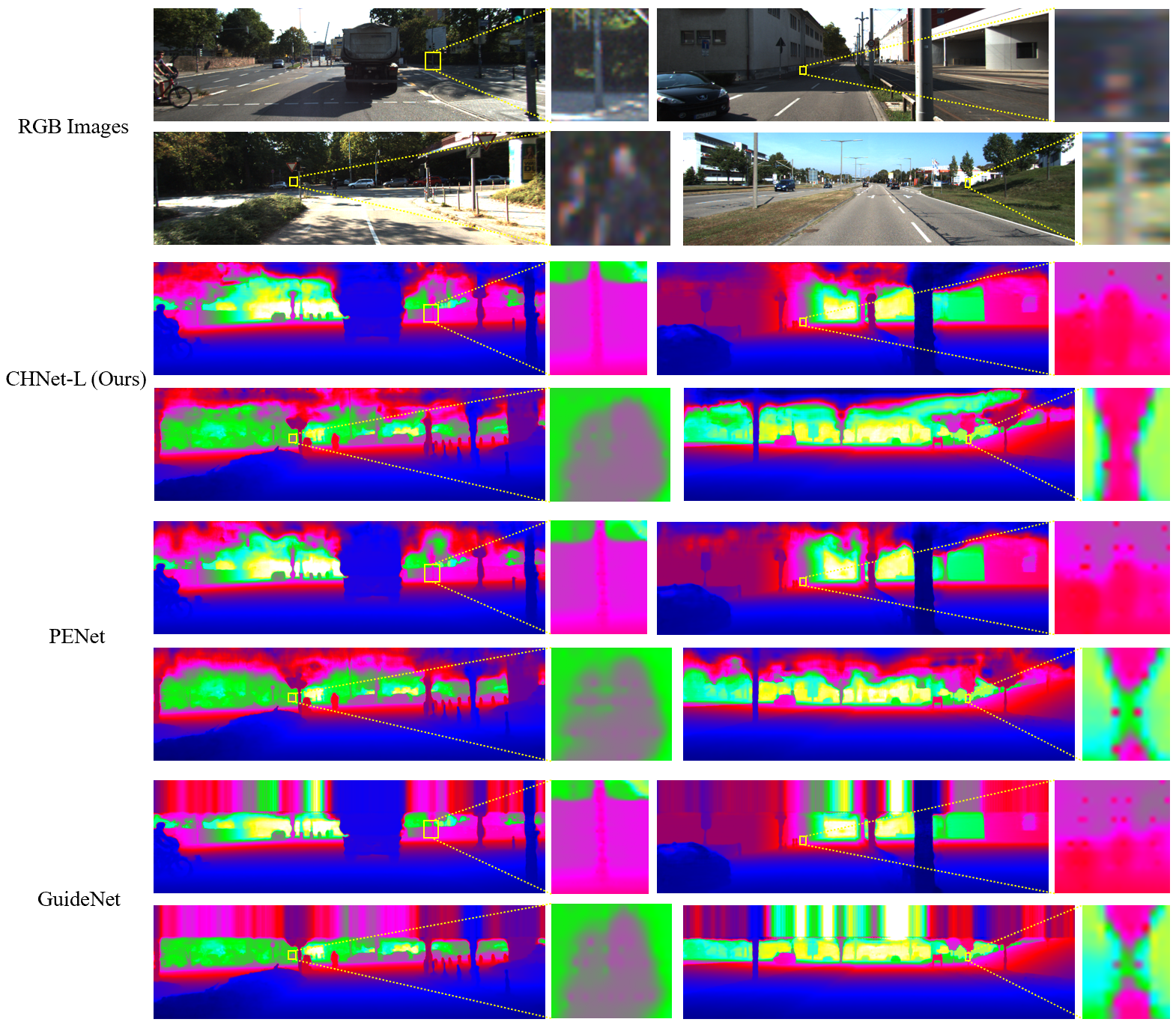}
  \caption{The prediction comparison of our CHNet-L and other methods on KITTI test set. The results are from the KITTI depth completion leaderboard in which depth images are colorized along with depth range. We also provide the RGB images for reference.}
  \label{fig:test-vis}
\end{figure*}

To verify the efficiency of our fast guidance module, we randomly generate a tensor with size (2, 128, 80, 304). This tensor is input into our fast guidance module and guide module in~\cite{9286883}, respectively, and the complexity comparison is shown in Table~\ref{tab:guidance_comparison}. Note that the module in~\cite{9286883} has been optimized by CUDA operators, but the proposed fast guidance module is still more than twice as fast. We also compare the performance under different density ratios of the input sparse depth map, as shown in Fig.~\ref{fig:DR}. The tested model is trained on the original train dataset, and we sample the validation dataset with density ratios of 0.2, 0.4, 0.6, and 0.8. As the number of valid depths decreases, the RMSE of all methods drops greatly. However, our CHNet-S still outperforms other models, which can prove that our method is more robust to the change of LiDAR density. To qualitatively compare the prediction details of our method and other models, we select typical examples from the KITTI depth completion leaderboard, as shown in Fig.~\ref{fig:test-vis}. Our CHNet-L has the ability to recover finer boundaries and more smooth details, as well as less noise in many scenarios.

\begin{figure*}[!t]
  \centering
  \includegraphics[width=0.9\linewidth]{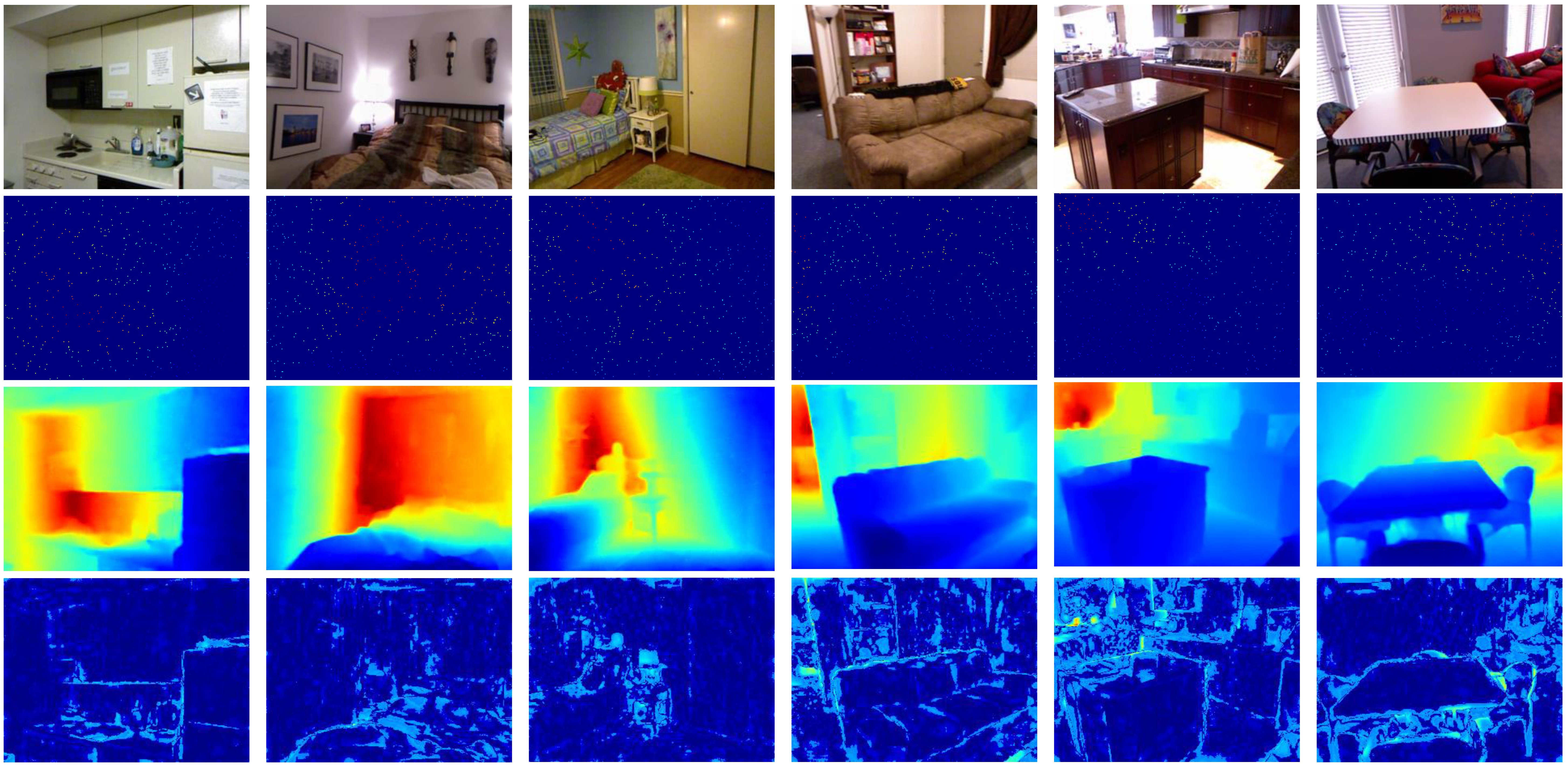}
  \caption{The visualization predictions of our method on NYUv2 dataset under different scenarios. The first row are the RGB image, the second is the sampled sparse depth map, the third is the depth map predicted by our method, and the fourth row is error between prediction and ground truth.}
  \label{fig:test-vis-nyu}
\end{figure*}

We also test our method on NYUv2 indoor dataset, and the comparison results are shown in Table~\ref{tab:sota-NYU}. Compared with the KITTI benchmark, the scale of NYUv2 dataset is smaller, so the $\delta_{1.25}$, $\delta_{1.25^2}$, and $\delta_{1.25^3}$ are similar among the state-of-the-art methods. Our CHNet achieves best result on RMSE, and slightly worse than GuideNet on REL. Although NYUv2 is from the indoor scenes, the results still can reflect the depth completion performance of our network, and the visualization predictions are shown in Fig.~\ref{fig:test-vis-nyu}.

\subsection{Ablation Studies}
In this section, we present the results of the ablation studies conducted to analyze the impact of various components on the performance of our proposed model. Note that we use the CHNet-S as the baseline.

First, we try to change the fusion strategy for sparse depth branch and image branch, and the comparison results are shown in Table~\ref{tab:fusion-strategy}. Simple summation and concatenation are coarse and obtain worse accuracy. We also compare our fast guidance module with the guide module in~\cite{9286883}, and our method achieves better result. Based on Table~\ref{tab:guidance_comparison} and Table~\ref{tab:fusion-strategy}, we can conclude that our guidance module is superior in terms of effectiveness and efficiency compared with that of GuideNet~\cite{9286883}.

\begin{table}[!t]
	\renewcommand{\arraystretch}{1.4}
	\renewcommand\tabcolsep{1pt}
	\caption{Ablation study for the fusion strategy.}
	\centering
	\label{tab:fusion-strategy}
	\begin{tabular}{c|cccc}
		\toprule
		Fusion Strategy                 & \textbf{RMSE}~$\downarrow$ & MAE~$\downarrow$ & iRMSE~$\downarrow$ & iMAE~$\downarrow$\\
		\midrule
        Summation                       & 1487.34   & 588.22  & 7.56   & 3.73  \\
        Concatenation                   & 1421.41   & 531.34  & 6.52   & 3.07  \\
        Guide Module in~\cite{9286883}  & 769.30    & 215.69  & 3.17   & 0.97  \\
        Fast guidance module (Ours)     & 763.87    & 212.08  & 2.38   & 0.94  \\
		\bottomrule
	\end{tabular}
\end{table}

\begin{table}[!t]
	\renewcommand{\arraystretch}{1.4}
	\renewcommand\tabcolsep{4pt}
	\caption{Ablation study for the setting of expansion ratio.}
	\centering
	\label{tab:ablation-in-er}
	\begin{tabular}{c|cccc}
		\toprule
		Expansion Ratio    & \textbf{RMSE}~$\downarrow$ & MAE~$\downarrow$ & iRMSE~$\downarrow$ & iMAE~$\downarrow$ \\
		\midrule
        1                  & 769.77    & 215.63  & 2.63   & 0.96  \\
        2                  & 768.85    & 213.25  & 2.87   & 0.94  \\
        3                  & 763.87    & 212.08  & 2.38   & 0.94  \\
		\bottomrule
	\end{tabular}
\end{table}

\begin{figure*}[!t]
  \centering
  \includegraphics[width=6.5in]{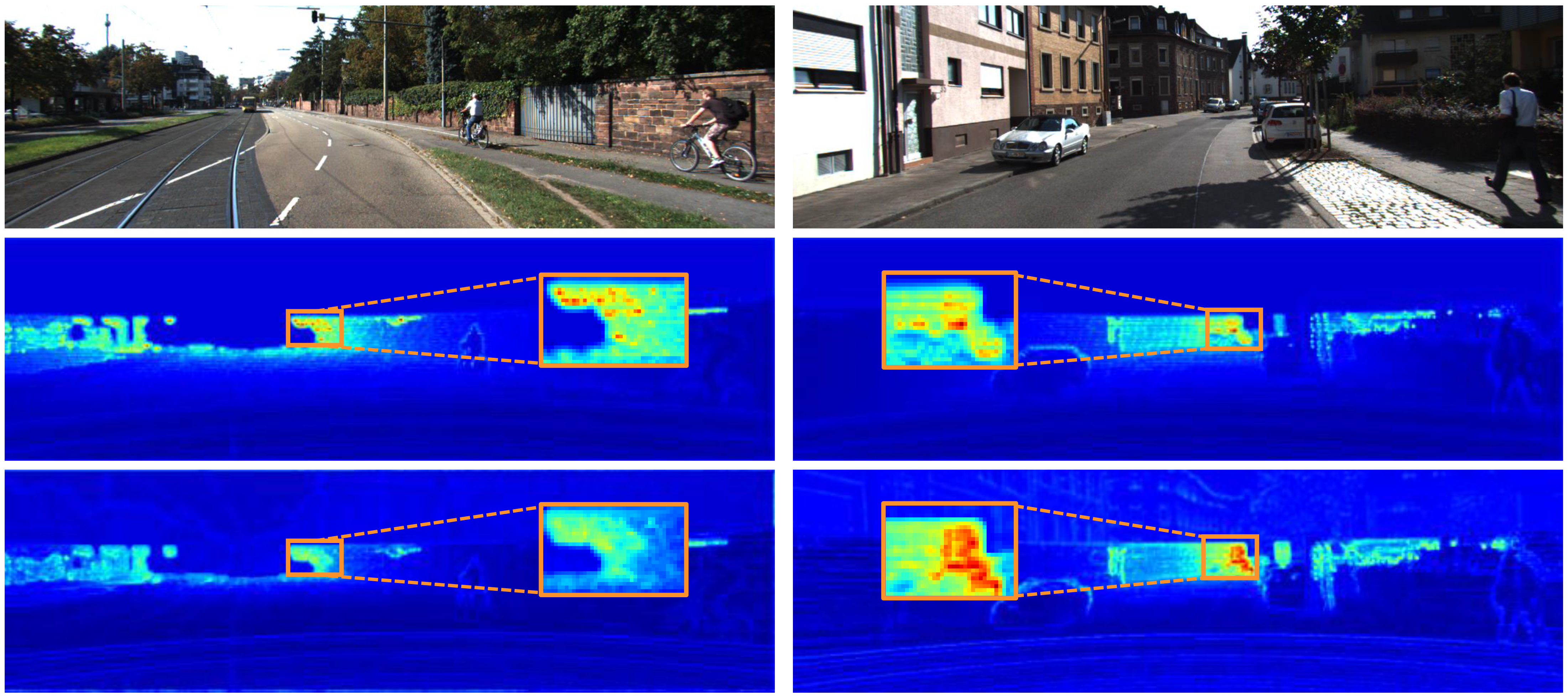}
  \caption{The effectiveness of our fast guidance module. The first row is the original image, and the second and third rows are the feature images before and after being processed by the fast guide module, respectively. Through the guidance module, more accurate outputs are obtained in the same region that should have similar depth values.}
  \label{fig:guide-effect}
\end{figure*}

\begin{figure*}[!t]
  \centering
  \includegraphics[width=6.5in]{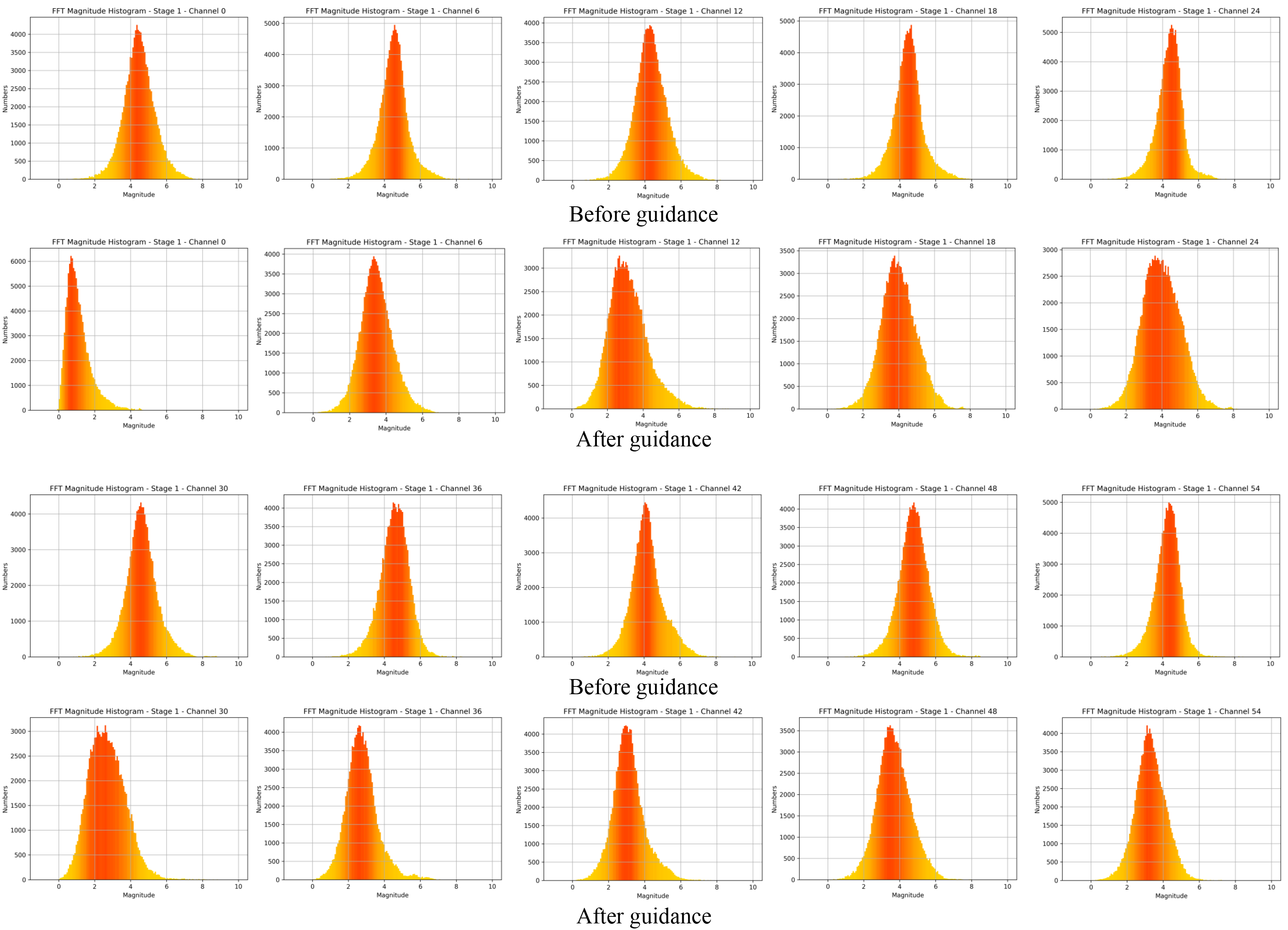}
  \caption{The frequency distribution histograms for feature maps of depth branch before and after the first guidance module. All selected ten channels represent more low-frequency signals that are produced through the RGB image guidance.}
  \label{fig:fq}
\end{figure*}

We also analyze the impact of the expansion ratio of our fast guidance module. As shown in Table~\ref{tab:ablation-in-er}, we evaluate the performance of the model for three different expansion ratios, i.e., 1, 2, and 3. From the results, we can see that the model achieves the best prediction with an expansion ratio of 3. This suggests that increasing the expansion ratio will improve the effectiveness of the model. We attribute this to the semantic richness of more subspace generation. The expansion ratio is not increased further for training speed, but the results show that there is still potential for further improvement.

To investigate the efficacy of our fast guidance module design, we conduct an ablation study focusing on its channel-wise multiplication (CW) and cross-channel multiplication (CC), as outlined in Table~\ref{tab:ablation-in-fgm}. Notably, the removal of CW leads to a significant performance deterioration, underscoring the pivotal role of the guidance kernel. Similarly, discarding CC also yields adverse effects on model performance. Subsequently, as shown in Table~\ref{tab:ablation-in-cca}, we evaluate the model's performance with different cross-channel aggregation strategies in the fast guidance module. None represents not using cross-channel aggregation, and it works worst among all metrics. Max means taking the maximum value at the corresponding position across the channel, while mean is average value. The results indicate that aggregating the information from different channels using the mean strategy is more effective.

\begin{table}[!t]
	\renewcommand{\arraystretch}{1.4}
	\renewcommand\tabcolsep{4pt}
    \caption{Ablation study for fast guidance module. CC is cross-channel multiplication, and CW is channel-wise multiplication.}
    \centering
    \begin{tabular}{cc|c|c|c|c}
    \toprule
        CW & CC & \textbf{RMSE}~$\downarrow$ & MAE~$\downarrow$ & iRMSE~$\downarrow$ & iMAE~$\downarrow$ \\ \hline
        \checkmark & \ding{55} & 776.49    & 214.69   & 2.59    & 0.96 \\
        \ding{55}& \checkmark & 791.28    & 218.23   & 2.50    & 0.97 \\
        \checkmark& \checkmark & 763.87    & 212.08   & 2.38    & 0.94  \\
    \bottomrule
    \end{tabular}
    \label{tab:ablation-in-fgm}
\end{table}

\begin{table}[!t]
	\renewcommand{\arraystretch}{1.4}
	\renewcommand\tabcolsep{4pt}
	\caption{Ablation study for the strategy of cross-channel aggregation.}
	\centering
	\label{tab:ablation-in-cca}
	\begin{tabular}{c|cccc}
		\toprule
		              & \textbf{RMSE}~$\downarrow$ & MAE~$\downarrow$ & iRMSE~$\downarrow$ & iMAE~$\downarrow$ \\
		\midrule
        None          & 776.49    & 214.69   & 2.59    & 0.96  \\
        Max           & 772.26    & 213.38   & 2.59    & 0.95  \\
        Mean          & 763.87    & 212.08   & 2.38    & 0.94  \\
		\bottomrule
	\end{tabular}
\end{table}

\begin{table}[!t]
	\renewcommand{\arraystretch}{1.4}
	\renewcommand\tabcolsep{4pt}
	\caption{RMSE comparison between different structures of prediction Head.}
	\centering
	\label{tab:ablation-in-ph}
	\begin{tabular}{c|cc}
		\toprule
		                    & Coupled Head   & Decoupled Head \\
		\midrule
        Unobserved Position & 804.97         & 798.71        \\
        Observed Position   & 532.87         & 505.78         \\
        Totality            & 772.17         & 763.87         \\
		\bottomrule
	\end{tabular}
\end{table}

\begin{table}[!t]
	\renewcommand{\arraystretch}{1.4}
	\renewcommand\tabcolsep{4pt}
	\caption{Complexity comparison between different structures of prediction Head.}
	\centering
	\label{tab:ablation-in-cph}
	\begin{tabular}{c|cc}
		\toprule
		                    & Coupled Head   & Decoupled Head \\
		\midrule
        Params              & 19K            & 38K        \\
        FLOPs               & 8G             & 16G         \\
        Speed               & 55ms           & 63ms         \\
		\bottomrule
	\end{tabular}
\end{table}

Finally, we examine different structures of the prediction head, as shown in Table~\ref{tab:ablation-in-ph}, which presents the results obtained by coupled and decoupled head structures. The decoupled head is superior to the coupled head in terms of RMSE for both observed and unobserved positions. This proves the correctness of our analysis for the optimization inconsistency problem and demonstrates that our decoupled head can alleviate it. The improvement of unobserved position is smaller than observed position. We think this is because the number of valid pixels is less, which makes the optimization of the unobserved position more favorable in the case of coupled head. To calculate the additional cost of the decoupled head, we compare the floating point operations (FLOPs) and parameters (Params) of the prediction head, and the inference time of the whole network, as shown in Table~\ref{tab:ablation-in-cph}. Based on the Table~\ref{tab:ablation-in-ph} and Table~\ref{tab:ablation-in-cph}, we can conclude that the decoupled head brings only an extra 8 ms of running time while achieving an 8.3 point improvement in RMSE.

\subsection{The Effectiveness of Fast Guidance Module}
In order to show the function of our fast guidance module more intuitively, we qualitatively show the feature map changes before and after the processing of the module, as shown in Fig.~\ref{fig:guide-effect}. When the intermediate features are input into our fast guidance module, more accurate outputs are obtained in the same region that should have similar depth values.

To analyze the principle of image guidance, we apply fast Fourier transform (FFT) to the feature maps of the LiDAR branch before and after the first guidance module, respectively, as shown in Fig.~\ref{fig:fq}. We select ten channels to make a comparison, and it is easily found that the low-frequency signal is enhanced after the guidance operation. This reflects that the semantic information representing different regions is injected into the depth feature, which helps the model perceive the layout of objects and scenes.

\section{CONCLUSIONS}
In this paper, we introduce CHNet, a concise but high-performing network tailored for depth completion. We offer two variants: CHNet-S prioritizes efficiency, while CHNet-L emphasizes effectiveness. Leveraging a compact structure and sophisticated modules, CHNet outperforms existing methods in terms of both prediction speed and accuracy across two diverse datasets. The fast guidance module enriches depth encoding by incorporating abundant semantic features within a local window, while the decoupled prediction head enables independent mapping for observed and unobserved positions.

In the future, we aim to pursue more fine-grained output while maintaining real-time performance. Additionally, we intend to delve deeper into the mechanisms through which RGB imagery contributes to depth learning, fostering further advancements in this domain.
\\
\\
\noindent{\textbf{CRediT authorship contribution statement}}
\\
\\
\textbf{Moyun Liu}: Conceptualization, Methodology, Formal analysis, Writing - original draft. \textbf{Bing Chen}: Formal analysis, Resources. \textbf{Youping Chen}: Supervision, Project administration. \textbf{Jingming Xie}: Investigation, Funding acquisition. \textbf{Yao Lei}: Writing - original draft, Data curation. \textbf{Yang Zhang}: Visualization, Validation. \textbf{Joey Tianyi Zhou}: Supervision, Methodology.
\\
\\
\noindent{\textbf{Declaration of Competing Interest}}
\\
\\
The authors declare that they have no known competing financial interests or personal relationships that could have appeared to influence the work reported in this paper.
\\
\\
\noindent{\textbf{Acknowledgement}}
\\
\\
This work is sponsored by the Hubei Provincial Key Research and Development Program Project of China (no. 2020BA B035). The financial support provided by China Scholarship Council Scholarship program (No.202206160061) is acknowledged.

%
%
%
%
%
%

\bibliographystyle{unsrt}

\begin{thebibliography}{10}

\bibitem{fernandes2021point}
Duarte Fernandes, Ant{\'o}nio Silva, Rafael N{\'e}voa, Cl{\'a}udia Sim{\~o}es,
  Dibet Gonzalez, Miguel Guevara, Paulo Novais, Jo{\~a}o Monteiro, and Pedro
  Melo-Pinto.
\newblock Point-cloud based 3d object detection and classification methods for
  self-driving applications: A survey and taxonomy.
\newblock {\em Information Fusion}, 68:161--191, 2021.

\bibitem{hu2022deep}
Junjie Hu, Chenyu Bao, Mete Ozay, Chenyou Fan, Qing Gao, Honghai Liu, and
  Tin~Lun Lam.
\newblock Deep depth completion from extremely sparse data: A survey.
\newblock {\em IEEE Transactions on Pattern Analysis and Machine Intelligence},
  pages 1--20, 2022.

\bibitem{zhou2023bcinet}
Wujie Zhou, Yuchun Yue, Meixin Fang, Xiaohong Qian, Rongwang Yang, and Lu~Yu.
\newblock Bcinet: Bilateral cross-modal interaction network for indoor scene
  understanding in rgb-d images.
\newblock {\em Information Fusion}, 94:32--42, 2023.

\bibitem{yang2023uplp}
Haozhi Yang, Jing Yuan, Yuanxi Gao, Xingyu Sun, and Xuebo Zhang.
\newblock Uplp-slam: Unified point-line-plane feature fusion for rgb-d visual
  slam.
\newblock {\em Information Fusion}, 96:51--65, 2023.

\bibitem{mosella20212d}
Albert Mosella-Montoro and Javier Ruiz-Hidalgo.
\newblock 2d--3d geometric fusion network using multi-neighbourhood graph
  convolution for rgb-d indoor scene classification.
\newblock {\em Information Fusion}, 76:46--54, 2021.

\bibitem{zhao2021adaptive}
Shanshan Zhao, Mingming Gong, Huan Fu, and Dacheng Tao.
\newblock Adaptive context-aware multi-modal network for depth completion.
\newblock {\em IEEE Transactions on Image Processing}, 30:5264--5276, 2021.

\bibitem{9286883}
Jie Tang, Fei-Peng Tian, Wei Feng, Jian Li, and Ping Tan.
\newblock Learning guided convolutional network for depth completion.
\newblock {\em IEEE Transactions on Image Processing}, 30:1116--1129, 2021.

\bibitem{yan2022rignet}
Zhiqiang Yan, Kun Wang, Xiang Li, Zhenyu Zhang, Jun Li, and Jian Yang.
\newblock Rignet: Repetitive image guided network for depth completion.
\newblock In {\em Proc. Eur. Conf. Comput. Vis. (ECCV)}, pages 214--230, Oct.
  2022.

\bibitem{9206740}
Yongchi Zhang, Ping Wei, Huan Li, and Nanning Zheng.
\newblock Multiscale adaptation fusion networks for depth completion.
\newblock In {\em Proc. IEEE Int. Joint Conf. Neural Netw. (IJCNN)}, pages
  1--7, Jul. 2020.

\bibitem{6319316}
Kaiming He, Jian Sun, and Xiaoou Tang.
\newblock Guided image filtering.
\newblock {\em IEEE Transactions on Pattern Analysis and Machine Intelligence},
  35(6):1397--1409, 2013.

\bibitem{cheng2018depth}
Xinjing Cheng, Peng Wang, and Ruigang Yang.
\newblock Depth estimation via affinity learned with convolutional spatial
  propagation network.
\newblock In {\em Proc. Eur. Conf. Comput. Vis. (ECCV)}, pages 103--119, Sep.
  2018.

\bibitem{Qiu2019deeplidar}
Jiaxiong Qiu, Zhaopeng Cui, Yinda Zhang, Xingdi Zhang, Shuaicheng Liu, Bing
  Zeng, and Marc Pollefeys.
\newblock Deeplidar: Deep surface normal guided depth prediction for outdoor
  scene from sparse lidar data and single color image.
\newblock In {\em Proc. IEEE Conf. Comput. Vis. Pattern Recognit. (CVPR)},
  pages 3313--3322, Jun. 2019.

\bibitem{hu2021penet}
Mu~Hu, Shuling Wang, Bin Li, Shiyu Ning, Li~Fan, and Xiaojin Gong.
\newblock Penet: Towards precise and efficient image guided depth completion.
\newblock In {\em Proc. IEEE Int. Conf. Robot. Automat. (ICRA)}, pages
  13656--13662, May. 2021.

\bibitem{uhrig2017sparsity}
Jonas Uhrig, Nick Schneider, Lukas Schneider, Uwe Franke, Thomas Brox, and
  Andreas Geiger.
\newblock Sparsity invariant cnns.
\newblock In {\em Proc. Int. Conf. on 3D Visi. (3DV)}, pages 11--20, Oct. 2017.

\bibitem{chen2022guided}
Long Chen and Qing Li.
\newblock Guided spatial propagation network for depth completion.
\newblock {\em IEEE Robotics and Automation Letters}, 7(4):12608--12614, 2022.

\bibitem{el2022guided}
Mohammad~Z El-Yabroudi, Ikhlas Abdel-Qader, Bradley~J Bazuin, Osama Abudayyeh,
  and Rakan~C Chabaan.
\newblock Guided depth completion with instance segmentation fusion in
  autonomous driving applications.
\newblock {\em Sensors}, 22(24):9578, 2022.

\bibitem{nazir2022semattnet}
Danish Nazir, Alain Pagani, Marcus Liwicki, Didier Stricker, and
  Muhammad~Zeshan Afzal.
\newblock Semattnet: Toward attention-based semantic aware guided depth
  completion.
\newblock {\em IEEE Access}, 10:120781--120791, 2022.

\bibitem{eldesokey2018propagating}
Abdelrahman Eldesokey, Michael Felsberg, and Fahad~Shahbaz Khan.
\newblock Propagating confidences through cnns for sparse data regression.
\newblock {\em arXiv preprint arXiv:1805.11913}, 2018.

\bibitem{jaritz2018sparse}
Maximilian Jaritz, Raoul De~Charette, Emilie Wirbel, Xavier Perrotton, and
  Fawzi Nashashibi.
\newblock Sparse and dense data with cnns: Depth completion and semantic
  segmentation.
\newblock In {\em Proc. Int. Conf. on 3D Visi. (3DV)}, pages 52--60, Sep. 2018.

\bibitem{knutsson1993normalized}
Hans Knutsson and C-F Westin.
\newblock Normalized and differential convolution.
\newblock In {\em Proc. IEEE Conf. Comput. Vis. Pattern Recognit. (CVPR)},
  pages 515--523, Jun. 1993.

\bibitem{glorot2011deep}
Xavier Glorot, Antoine Bordes, and Yoshua Bengio.
\newblock Deep sparse rectifier neural networks.
\newblock In {\em Proc. Int. Conf. Artif. Intell. Statist. (AISTATS)}, pages
  315--323, Apr. 2011.

\bibitem{lu2020depth}
Kaiyue Lu, Nick Barnes, Saeed Anwar, and Liang Zheng.
\newblock From depth what can you see? depth completion via auxiliary image
  reconstruction.
\newblock In {\em Proc. IEEE Conf. Comput. Vis. Pattern Recognit. (CVPR)},
  pages 11306--11315, Jun. 2020.

\bibitem{yu2021grayscale}
Qingyang Yu, Lei Chu, Qi~Wu, and Ling Pei.
\newblock Grayscale and normal guided depth completion with a low-cost lidar.
\newblock In {\em Proc. IEEE Int. Conf. Inf. Process. (ICIP)}, pages 979--983,
  Sep. 2021.

\bibitem{li2020multi}
Ang Li, Zejian Yuan, Yonggen Ling, Wanchao Chi, Chong Zhang, et~al.
\newblock A multi-scale guided cascade hourglass network for depth completion.
\newblock In {\em Proc. IEEE Winter Conf. Appli. of Comput. Vis. (WACV)}, pages
  32--40, Mar. 2020.

\bibitem{liu2021fcfr}
Lina Liu, Xibin Song, Xiaoyang Lyu, Junwei Diao, Mengmeng Wang, Yong Liu, and
  Liangjun Zhang.
\newblock Fcfr-net: Feature fusion based coarse-to-fine residual learning for
  depth completion.
\newblock In {\em Proc. AAAI Conf. Artif. Intell. (AAAI)}, volume~35, pages
  2136--2144, Feb. 2021.

\bibitem{MaCK19}
Fangchang Ma, Guilherme~Venturelli Cavalheiro, and Sertac Karaman.
\newblock Self-supervised sparse-to-dense: Self-supervised depth completion
  from lidar and monocular camera.
\newblock In {\em Proc. IEEE Int. Conf. Robot. Automat. (ICRA)}, pages
  3288--3295, May. 2019.

\bibitem{dimitrievski2018learning}
Martin Dimitrievski, Peter Veelaert, and Wilfried Philips.
\newblock Learning morphological operators for depth completion.
\newblock In {\em Proc. Int. Conf. Adv. Concepts Intell. Vis. Syst. (ACIVS)},
  pages 450--461, Sep. 2018.

\bibitem{chen2020dynamic}
Yinpeng Chen, Xiyang Dai, Mengchen Liu, Dongdong Chen, Lu~Yuan, and Zicheng
  Liu.
\newblock Dynamic convolution: Attention over convolution kernels.
\newblock In {\em Proc. IEEE Conf. Comput. Vis. Pattern Recognit. (CVPR)},
  pages 11030--11039, Jun. 2020.

\bibitem{zhang2023completionformer}
Youmin Zhang, Xianda Guo, Matteo Poggi, Zheng Zhu, Guan Huang, and Stefano
  Mattoccia.
\newblock Completionformer: Depth completion with convolutions and vision
  transformers.
\newblock In {\em Proc. IEEE Conf. Comput. Vis. Pattern Recognit. (CVPR)},
  pages 18527--18536, 2023.

\bibitem{cheng2020cspn++}
Xinjing Cheng, Peng Wang, Chenye Guan, and Ruigang Yang.
\newblock Cspn++: Learning context and resource aware convolutional spatial
  propagation networks for depth completion.
\newblock In {\em Proc. AAAI Conf. Artif. Intell. (AAAI)}, volume~34, pages
  10615--10622, Feb. 2020.

\bibitem{he2016deep}
Kaiming He, Xiangyu Zhang, Shaoqing Ren, and Jian Sun.
\newblock Deep residual learning for image recognition.
\newblock In {\em Proc. IEEE Conf. Comput. Vis. Pattern Recognit. (CVPR)},
  pages 770--778, Jun. 2016.

\bibitem{jung2021fine}
Hyunyoung Jung, Eunhyeok Park, and Sungjoo Yoo.
\newblock Fine-grained semantics-aware representation enhancement for
  self-supervised monocular depth estimation.
\newblock In {\em Proc. IEEE Int. Conf. Comput. Vis. (ICCV)}, pages
  12642--12652, 2021.

\bibitem{dai2020parameters}
Dawei Dai, Liping Yu, and Hui Wei.
\newblock Parameters sharing in residual neural networks.
\newblock {\em Neural Processing Letters}, 51:1393--1410, 2020.

\bibitem{silberman2012indoor}
Nathan Silberman, Derek Hoiem, Pushmeet Kohli, and Rob Fergus.
\newblock Indoor segmentation and support inference from rgbd images.
\newblock In {\em Proc. Eur. Conf. Comput. Vis. (ECCV)}, pages 746--760.
  Springer, 2012.

\bibitem{kingma2014adam}
Diederik~P Kingma and Jimmy Ba.
\newblock Adam: A method for stochastic optimization.
\newblock {\em arXiv preprint arXiv:1412.6980}, 2014.

\bibitem{xu2019depth}
Yan Xu, Xinge Zhu, Jianping Shi, Guofeng Zhang, Hujun Bao, and Hongsheng Li.
\newblock Depth completion from sparse lidar data with depth-normal
  constraints.
\newblock In {\em Proc. IEEE Int. Conf. Comput. Vis. (ICCV)}, pages 2811--2820,
  Oct. 2019.

\bibitem{chen2022depth}
Hu~Chen, Hongyu Yang, Yi~Zhang, et~al.
\newblock Depth completion using geometry-aware embedding.
\newblock In {\em 2022 International Conference on Robotics and Automation
  (ICRA)}, pages 8680--8686. IEEE, 2022.

\bibitem{xu2020deformable}
Zheyuan Xu, Hongche Yin, and Jian Yao.
\newblock Deformable spatial propagation networks for depth completion.
\newblock In {\em Proc. IEEE Int. Conf. Inf. Process. (ICIP)}, pages 913--917,
  Nov. 2020.

\bibitem{jeon2021abcd}
Yurim Jeon, Hwichang Kim, and Seung-Woo Seo.
\newblock Abcd: Attentive bilateral convolutional network for robust depth
  completion.
\newblock {\em IEEE Robotics and Automation Letters}, 7(1):81--87, 2021.

\bibitem{chen2019learning}
Yun Chen, Bin Yang, Ming Liang, and Raquel Urtasun.
\newblock Learning joint 2d-3d representations for depth completion.
\newblock In {\em Proc. IEEE Int. Conf. Comput. Vis. (ICCV)}, pages
  10023--10032, Oct. 2019.

\bibitem{fan2022cascade}
Rizhao Fan, Zhigen Li, Matteo Poggi, and Stefano Mattoccia.
\newblock A cascade dense connection fusion network for depth completion.
\newblock In {\em The 33rd British Machine Vision Conference}, volume~1,
  page~2, 2022.

\bibitem{ke2021mdanet}
Yanjie Ke, Kun Li, Wei Yang, Zhenbo Xu, Dayang Hao, Liusheng Huang, and Gang
  Wang.
\newblock Mdanet: Multi-modal deep aggregation network for depth completion.
\newblock In {\em Proc. IEEE Int. Conf. Robot. Automat. (ICRA)}, pages
  4288--4294, May. 2021.

\bibitem{sun2023revisiting}
Xinglong Sun, Jean Ponce, and Yu-Xiong Wang.
\newblock Revisiting deformable convolution for depth completion.
\newblock In {\em 2023 IEEE/RSJ International Conference on Intelligent Robots
  and Systems (IROS)}, pages 1300--1306. IEEE, 2023.

\bibitem{liu2023mff}
Lina Liu, Xibin Song, Jiadai Sun, Xiaoyang Lyu, Lin Li, Yong Liu, and Liangjun
  Zhang.
\newblock Mff-net: Towards efficient monocular depth completion with
  multi-modal feature fusion.
\newblock {\em IEEE Robotics and Automation Letters}, 8(2):920--927, 2023.

\bibitem{zhang2018deep}
Yinda Zhang and Thomas Funkhouser.
\newblock Deep depth completion of a single rgb-d image.
\newblock In {\em Proc. IEEE Conf. Comput. Vis. Pattern Recognit. (CVPR)},
  pages 175--185, 2018.

\bibitem{imran2019depth}
Saif Imran, Yunfei Long, Xiaoming Liu, and Daniel Morris.
\newblock Depth coefficients for depth completion.
\newblock In {\em Proc. IEEE Conf. Comput. Vis. Pattern Recognit. (CVPR)},
  pages 12438--12447, Jun. 2019.

\bibitem{lee2021depth}
Byeong-Uk Lee, Kyunghyun Lee, and In~So Kweon.
\newblock Depth completion using plane-residual representation.
\newblock In {\em Proc. IEEE Conf. Comput. Vis. Pattern Recognit. (CVPR)},
  pages 13916--13925, 2021.

\end{thebibliography}

\end{document}